\documentclass[runningheads]{llncs}
\usepackage[T1]{fontenc}
\usepackage{graphicx}
\usepackage{mwe}
\usepackage{todonotes}
\usepackage{subcaption}
\usepackage{multirow}
\usepackage{hyperref}
\usepackage{bbding}
\usepackage{color}

\newcommand{\added}[1]{\textcolor{black}{#1}}
\begin{document}
\title{From Development to Deployment of AI-assisted Telehealth and Screening for Vision- and Hearing-threatening diseases in resource-constrained settings: Field Observations, Challenges and Way Forward}
%
%
\author{Mahesh Shakya\inst{1}\orcidID{0009-0006-3784-9608}\Envelope \and
Bijay Adhikari\inst{2}\orcidID{0009-0005-4503-783X} \and
Nirsara Shrestha\inst{2}\orcidID{0000-0001-6052-1201} \and
Bipin Koirala\inst{2}\orcidID{0000-0001-9108-4825}\and
Arun Adhikari\inst{2}\orcidID{0000-0002-8831-6194}\and
Prasanta Poudyal\inst{2}\orcidID{0000-0003-4645-852X}\and
Luna Mathema\inst{2}\orcidID{0000-0001-8493-3315} \and
Sarbagya Buddhacharya\inst{2}\orcidID{0000-0002-3678-0083} \and
Bijay Khatri\inst{2}\orcidID{0000-0001-9727-9877} \and
Bishesh Khanal\inst{1}\orcidID{0000-0002-2775-4748}}
%
\authorrunning{M. Shakya et al.}
%
\institute{Nepal Applied Mathematics and Informatics Institute for research (NAAMII), Lalitpur, Nepal \and
Hospital for Children, Eye, ENT and Rehabilitation Services (CHEERS), BP Eye Foundation, Bhaktapur, Nepal \\
\email{\{mahesh.shakya,bishesh.khanal\}@naamii.org.np}\\\email{\{bijayadhikari.bpef,nirsara.shrestha,lunashrestham,\\sarbagya.budhhacharya,bijay.bpef\}@gmail.com}}

%
\maketitle              
\begin{abstract}

Vision- and hearing-threatening diseases cause preventable disability, especially in resource-constrained settings(RCS) with few specialists and limited screening setup. 
Large scale AI-assisted screening and telehealth has potential to expand early detection, but practical deployment is challenging in paper-based workflows and limited documented field experience exist to build upon.

We provide insights on challenges and ways forward in development to adoption of scalable AI-assisted Telehealth and screening in such settings.
Specifically, we find that iterative, interdisciplinary collaboration through early prototyping, shadow deployment and continuous feedback is important to build shared understanding as well as reduce usability hurdles when transitioning from paper-based to AI-ready workflows.
We find public datasets and AI models highly useful despite poor performance due to domain shift.
In addition, we find the need for automated AI-based image quality check to capture gradable images for robust screening in high-volume camps.

Our field learning stress the importance of treating AI development and workflow digitization as an end-to-end, iterative co-design process.
By documenting these practical challenges and lessons learned, we aim to address the gap in contextual, actionable field knowledge for building real-world AI-assisted telehealth and mass-screening programs in RCS.

\end{abstract}
\section{Introduction}
Vision loss from conditions such as Diabetic Retinopathy (DR), Glaucoma, and Age-related Macular Degeneration (AMD) as well as preventable hearing loss due to middle ear diseases like Otitis Media is often avoidable if these conditions are detected early through systematic screening\cite{singer1992screening,topouzis2007glaucoma,bressler2002early,rosenfeld2016clinical}. 
Early identification and timely intervention can significantly reduce the risk of irreversible vision or hearing impairment, especially in settings where patients otherwise present late in the disease course.

However, resource-constrained settings lack ophthalmologists and otolaryngologists for diagnosis and treatment of such conditions, and do not have robust, structured screening infrastructure\cite{burton2021lancet,piyasena2019systematic}. 
Such settings may consist of Primary Care Centers (PCC) where community health workers (CHW) screen and treat common eye and ear conditions and are the first point of contact for health services for large section of the population in Low and Middle Income Countries(LMICs). These centers are often linked with Hub Hospitals providing specialist services and act as referral destination for these centers\cite{srivastava2020development}.
Basic tools (often analog) such as ophthalmoscope, slit lamp, fundus camera, audiometer, otoscope etc. are available to observe structural abnormalities but lack advanced confirmatory tools like visual field analyzers, optical coherence tomography (OCT), or audiometry booths and required expertise for comprehensive vision and hearing tests. 

While these extended services help deliver care beyond urban hospitals and yearly short-term trainings are provided to improve screening capabilities, they are often constrained by limited operator skill and varying levels of diagnostic expertise in screening\cite{sabri2019paediatric,ramkumar2019implementation}.

Artificial Intelligence (AI) offers a promising solution to bridge these capacity gaps\cite{ciecierski2022artificial}. 
When embedded within telehealth and screening workflows, AI could assist non-specialist health workers in rapidly identifying referable cases as well as assist clinicians in triaging referred cases, expanding screening reach, enabling timely referrals to higher-level centers and scaling services. 
A possible model may involve AI model flagging suspected referable cases, which are then reviewed remotely by specialists at a tertiary hospital, creating a tiered, efficient workflow.

Despite these opportunities, developing and deploying AI models at the point of care is still highly challenging-especially in low-resource\cite{krones2024theoretical,cabitza2020bridging,ciecierski2022artificial} settings where workflow remain paper-based and use of analog image taking devices are common (For e.g. analog instead of digital otoscopes, analog slit lamp instead of digital portable fundus camera) with limited experience of Healthcare Professionals in AI-based Projects and that of AI researchers limited experience outside of lab settings resulting in trust barriers and terminology gaps.
There is a need to co-design and mutual learning among interdisciplinary teams towards shared understanding for meaningful AI-assisted workflow that health workers can own up.


\begin{figure}
    \centering
    \includegraphics[width=1.0\linewidth]{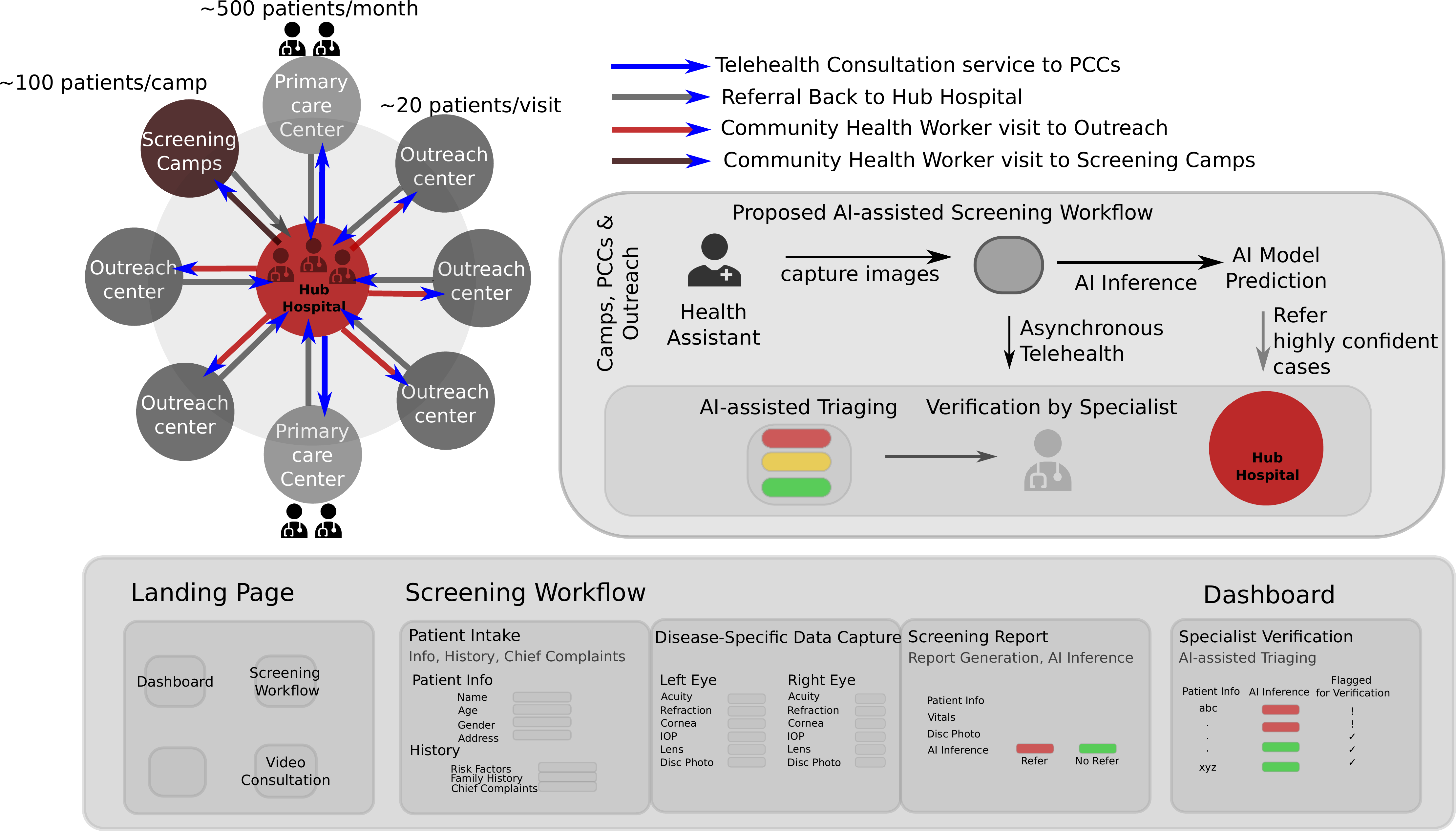}
    \caption{\textit{Left}:Hub-and-Spoke Model of Healthcare Delivery which represents our current clinical context. \textit{Right \& Bottom}: Prototype AI-assisted Screening, Telehealth and Referral Workflow and WebApp being tested in the current study.}
    \label{fig:hub_spoke}
\end{figure}

We started AI-assisted Telehealth Project for the very first time at the Hospital site with the mandate to apply AI to improve service delivery to PCCs, outreach and screening camps in telehealth setting. Except the Lead Computer Scientist, this was the first time being involved in building a real world AI-based workflow from scratch for most of the members of the Clinical team as well as the AI Researchers. In this unique scenario of multidisciplinary collaboration in AI-powered Telehealth in RCS, useful insights were gained with limited documentation available in the literature. 

Hence, this paper contributes practical experience and documentation working toward AI-ready and AI-assisted screening and telehealth workflows for vision- and hearing-threatening diseases in this unique context. Specifically, We contribute
\begin{enumerate}
    \item Field experience from screening camps and outreach centers highlighting practical usability issues such as time constraints, manual image transfer, intake form burdens requiring stakeholder-informed workflow design
    \item Quantification of image acquisition challenges from digital otoscopes in high throughput screening camps and need for automated image quality assessment to support consistent, reliable screening
    \item Observations on how public datasets and AI models while often showing reduced performance in new real-world settings, are highly valuable for workflow prototyping and building shared understanding between AI teams and clinicians 
\end{enumerate}

\section{Related Work}
Several studies have explored real-world AI deployment for screening vision- and hearing-threatening diseases in both well-resourced\cite{li2021proposed,skevas2024implementing} and resource-constrained settings\cite{sarmiento2025guiding,yim2021barriers,rani2025smart}.
Studies in well-resourced settings have evaluated structured ML-aided triaging and grading framework showing potential benefits in prioritization and saving consultation time\cite{li2021proposed,skevas2024implementing}.

On the other hand, evidences from resource-constrained settings remain scarce. 
\cite{bellemo2019artificial} validated a deep learning model for diabetic retinopathy screening in Zambian mobile clinics using Singaporean training data demonstrating good predictive performance in a new context.
But these studies do not address challenges such as workflow integration and interdisciplinary collaboration required for translating from clinical gaps to AI problem formulation.

\added{Additionally, various works\cite{mollura2020artificial,okolo2020ai,weber2010remembering,elahi2020overcoming}, including reviews\cite{cabitza2020bridging,ciecierski2022artificial}, have highlighted unique challenges and nuances of deployment in LMICs such as need for digital infrastructure, phased introduction of AI\cite{mollura2020artificial}, inclusive design\cite{okolo2020ai}, digitization of paper forms\cite{singh2009numeric}.}



Together, these works show that while AI-assisted screening and telehealth is feasible, much of the literature still focuses on evaluation of AI model performance and pilot studies rather than the full development-to-deployment cycle, including experiences in interdisciplinary collaboration, building trust when the team is exploring AI-powered systems in clinical workflow for the first time.
\added{Our work adds to the latter discussion on development to deployment challenges specifically in the hub-and-spoke model of healthcare delivery.}

\section{Methods}


First, we outline the context in which the project is being implemented. The initiative is led by the Ophthalmology and Otolaryngology Department of a tertiary-level hospital operating under a ``hub-and-spoke'' model. As shown in Figure~\ref{fig:hub_spoke}, this central hospital supports five outreach centers and two Primary Eye and Ear-Nose-Throat(ENT) Centers(PEEC) which act as ``spokes'' from which referrals are made back to the ``hub''. Health Workers from the central hub visit outreach centers weekly whereas CHWs are stationed permanently at PEECs. 

As a foundation for AI-ready workflow, we start by pre-testing/prototyping a Digitized Data Collection Workflow replicating the paper-based proforma documents used by Ophthalmic Assistants as well as ENT Assistants and then reiterating the collection workflow by field testing them in a single high-volume screening camp as well as one of the PEECs. 




Additionally, to evaluate the feasibility of AI-assisted screening workflow in high throughput setting, we use AI models trained on publicly available datasets whenever feasible in parallel with data collection. 
\textit{Fundus-based Ophthalmology Screening using AI Models trained on Public Datasets} - We use the publicly available EyePACS Glaucoma Justified Referral Dataset \cite{lemij2023characteristics} to train a lightweight classifier, MobileNetv3\cite{howard2019searching}, where the focus is on ability to make inferences on low-cost hardware such as standard mid-range laptop without GPU as well as a placeholder to describe the overall workflow to clinicians.
We trained the ImageNet1K-pretrained MobileNetv3 architecture\cite{howard2019searching} on fundus images square-cropped to 512x512 px. The model was trained till 10 epoch with Adam optimizer with learning rate $1e-3$ and batch size 4. Various Augmentations (Uniform Random Noise(0.01), Color Jitter(0.01), Random Horizontal Flipping) were done.

Additionally, we use the retrospective images stored in the Fundus Camera Devices since last 2 years in the Tertiary Hospital to test the robustness of the available multi-center large datasets and subsequent trained AI model performance in unseen context. This is described in the Datasets subsection below.
\textit{Otoscopy-based ENT Screening using AI Models trained on Public Dataset} - We find that the publicly available datasets for Otoscopy-based Screening lacked necessary metadata, low in quality as well as quantity resulting in unusable AI model. 
Additionally, retrospective otoscopic images from tertiary hospital were not available because of predominant use of analog otscopes in clinic and even when digital otoscopes were available, image storage was deemed too cumbersome or unnecessary.

Parallely, we also tested a web-based platform requiring internet connectivity to send the images to a on-premise server at the tertiary hospital and then run inference on the server and return the prediction to the user to evaluate feasibility of AI-assisted Triaging and Specialist Verification as shown in Figure~\ref{fig:hub_spoke}. This prototype system was shadow deployed in Primary Care Centers and screening camps as part of our workflow understanding, calibration and needs assessment phase.

\subsubsection{Datasets} 
\textit{Internal Test set - Mixed - EyePacs} We used a subset of the EyePACS AIROGS Dataset as described in \cite{steen2023standardized}. The dataset consists of balanced class labels (Referrable Glaucoma vs. Non Glaucoma) with 2500 images in train, and 385 images each in validation and test sets.

\textit{External Test set - Portable - Optomed AuroraIQ} This dataset consists of Fundus Images from 60 Patients taken with the portable Fundus Camera (specifically Optomed AuroraIQ). These images were taken from a Primary Care Center by an Ophthalmic Assistant. The images were taken in the context of real screening setting with multiple images of a single eye taken, if need be. These Images were labelled as Referrable Glaucoma vs. Non Glaucoma at the patient-level by the Glaucoma Specialist.

\textit{External Test set - Desktop - TopCon OCT1 Maestro and Canon CX1} The TopCon OCT1 Maestro and the Canon CX1 dataset consists of fundus images of patients suspected to have either a retinal disease or glaucoma by the clinician at the tertiary hospital, taken respectively from specific Desktop Fundus Cameras devices - TopCon OCT1 Maestro and Canon CX1. The distribution of this dataset consists of predominantly diseased fundus images.  We took 100 patient with fundus images with labelled Glaucoma at the image-level by the Specialist.

\section{Results}
\subsection{Field Experience in Workflow Digitization and Shadow Deployment}
During the field testing, we embedded the AI output alongside routine clinical image review without influencing referral decisions, enabling clinicians and technicians to assess its usability in parallel with routine practice. 
Usability observations revealed key bottlenecks:
\begin{enumerate}
    \item \textbf{Data Entry details} as staff spent considerable time filling paper or digital forms with patient demographics before image capture.
    \item \textbf{Manual transfer frictions}, where images had to be copied from the portable fundus camera via USB wire to the laptop running the web-based platform, often adding several minutes per patient.
    \item \textbf{User Feedback} Worried about the time needed to fill in the proforma document digitally. We initially had all the fields as required. User experience issues such as being able to quicken the form entry via autocompletes, drop-down menus were added after the first testing on-site at the screening camp.
    \item Despite these frictions, staff appreciate the \textbf{immediate feedback} on potential disease diagnosis.
    
\end{enumerate}
\textit{Recommendation} Despite many offerings available for digitization of the workflow, there are still barriers, one of which is perceived time consuming aspect of the digitization\cite{kabukye2023implementing}. Despite the web app we built, frontline healthworkers were still hesitant especially in high-volume patient flow. 
To allow filling of the proforma details hands-free while simultaneously catering to the patient via voice-to-text based automated data entry in such noisy environment may be good future direction.

\subsection{Quality Image Acquisition Challenges in Asynchronous Telehealth and Otoscopy-based Screening}
One of the barriers in deploying AI-assisted screening is inadequate image quality \cite{de2021ophthalmic,chen2023barriers}, which directly affects diagnostic reliability and AI model performance. In our pilot visits to Primary Care Centers and screening camps, as well as data collection site at tertiary center, we observed that non-trivial proportion of otoscopic images captured by health assistants were not of diagnostic quality, leading to failed remote grading. 

\textit{Otoscopic Image-based Grading}
To quantify the extent to which image quality issues occur when taken by Health Assistants for off-site consultation/screening in telehealth for ear screening, we conducted a small-scale image quality grading and screening for diseases of ear images collected with a digital otoscope in a field setting.
Four otology specialists independently classified each of 100 tympanic membrane images into one of four diagnostic categories - Normal, Acute Otitis Media (AOM), Otitis Media with Effusion (OME), Chronic Otitis Media (COM) - or selected ‘Not Possible to Determine (NPD)’ when the image quality or contextual information was insufficient. All specialists graded all images, enabling calculation of inter-rater reliability. 

As shown in Table~\ref{tab:annotation_otoscopic_images}, We find that significant portion of the otoscopic images taken during screening camps by frontline Health Workers are ``ungradable'' mostly either due to poor image quality or obstruction due to earwax or foreign bodies. This also resulted in weak interannotator agreement with Fleiss' Kappa score of 0.37.
\begin{table}[htpb]
    \centering
    \resizebox{0.4\columnwidth}{!}{%
    \begin{tabular}{|l|c|c|c|c|}
        \hline
         \multirow{2}{*}{Categories} & \multicolumn{4}{c|}{Clinicians} \\ \cline{2-5} 
         & A & B & C & D\\ \hline
         \textbf{Possible to Determine} &  \textbf{76} & \textbf{60} & \textbf{51} & \textbf{39} \\\hline
         Normal &  47 & 17 &  20& 15\\\hline
         OME &  3& 12 & 4 & 6\\\hline         
         AOM &  5& 8 & 2 & 7\\\hline
         COM &  21& 23 & 25  & 11\\\hline
         \textbf{Not Possible to Determine} & \textbf{26} & \textbf{40} & \textbf{55} & \textbf{62}\\\hline
         Poor Image Quality & 22 & 29 & 52 & 55\\\hline
         Obstruction & 11 & 20 & 9 & 23\\\hline
         Inadequate Patient History & 1 & 8 & 1 & 0\\\hline
         Hearing Assessment Required & 1 & 7 & 0 & 2\\\hline
    \end{tabular}
    }
    \caption{Annotation results for 100 otoscopic images reviewed independently by four clinicians (A–D). The results show the number of images classified as “Possible to Determine” with specific diagnostic labels (Normal, OME, AOM, COM) and those marked as “Not Possible to Determine” further broken down by reasons such as poor image quality, obstruction (e.g., cerumen or foreign bodies), inadequate patient history, or need for additional hearing assessment. }
    \label{tab:annotation_otoscopic_images}
\end{table}

Time pressure in high-throughput screening camps, having no time to clean or replace the speculum when contaminated with earwax, and in-general the difficulty of visualizing the full tympanic membrane in otoscopic images lead to ambiguity or unusable images for offsite diagnosis.

\textit{Recommendation} This underscores the need to integrate robust image quality assessment mechanism in both AI-assisted and asynchronous telehealth workflow through targeted training and automated AI-based feedback on gradability. 

\subsection{Publicly available Datasets and AI Models are highly useful for workflow design and interdisciplinary discussion}

We find AI models trained on publicly available datasets are highly valuable as a starting point for designing and testing feasibility of AI-assisted screening workflows. In our project, using a glaucoma screening model trained on the widely used EyePACS Dataset\cite{lemij2023characteristics,steen2023standardized} allowed us to introduce a concrete AI tool to clinical teams and frontline health workers who had no prior experience working with AI-enabled systems.

This exposure helped establish a shared understanding of how AI could be integrated into routine screening and telehealth workflows for flagging potential referable cases. By testing a working prototype, clinicians and outreach teams could see where and how AI might fit in - including how image capture, transfer and AI output could be embedded in existing screening and referral pathways. This facilitated interdisciplinary discussions, clarified expectations, and helped uncover usability and integration challenges early in the design process.

However, while these public models serve well for early-stage prototyping and workflow co-design, they do not perform reliably as drop-in solutions for local deployment. We show this by testing Zero-shot performance of an AI model trained on public dataset to classify Referrable Glaucoma(RG) vs. Non(Refereable) Glaucoma(NRG), testing their performance on datasets created from our screening camps and Outpatient Department at the tertiary hospital.

As shown in Table~\ref{tab:model performance on device-induced domain shift}, We find that although the model performs exceptionally well(>90\%) on it's internal test set (with similar population and imaging devices) across all classification metrics, it's zero-shot performance already drops to $\sim$84\% on portable fundus images taken from our screening camps. Moreover, the model's Precision reduces drastically to $\sim$57\% and $\sim$37\% respectively on images taken with TopCon OCT1 and Canon CX1 respectively. Various device-related artifacts may result in domain-shift and hence drop in performance.


\begin{table}
    \centering
    \begin{tabular}{|c|c|c|c|c|c|c|c|}
        \hline Dataset  & Test-set Type &   \#RG/NRG  & Accuracy & Precision & Recall & F1\\ \hline
        EyePacs & Internal & 385/385 & 0.9208 & 0.9219 & 0.9195 & 0.9207 \\ \hline
        Optomed AuroraIQ  & External & 28/32 &0.8435 & 0.8367 & 0.8039 & 0.8200 \\ \hline
        TopCon OCT1 Maestro & External & 100/0  & - & 0.5722 & - &  -\\ \hline
        Canon CX1 & External & 100/0  & - & 0.3370 & - & - \\ \hline        
    \end{tabular}
    \caption{\textit{Zero-shot performance of Glaucoma Screening AI model trained on a subset of EyePACS Dataset and tested on population and devices available in our setting}: The performance of the AI model on it's internal test set is shown for comparison against external test sets made from images captured from screening camps and Outpatient Department.}
    \label{tab:model performance on device-induced domain shift}
\end{table}


\section{Discussion}

Beyond image quality and usability, we faced challenges in translating clinical needs into robust AI tasks. Groundtruth definition is rarely straightforward, as diagnostic categories often hinge on subjective thresholds. For instance, unlike the EyePACS protocol that labels glaucoma suspects only at cup-to-disc ratios >0.7, our team used lower thresholds prioritizing over-referral since routine screening is rare in RCS. Such choices highlight the need to explicitly document protocols, local risk preferences, and diagnostic assumptions in AI screening tool development.
Additionally, the context-specific challenges of workflow digitization may require tailor-made solutions whereas challenges due to domain-shift and image quality may be broadly applicable. Advances such as Federated Learning, synthetic data augmentation techniques may benefit multisite data aggregation without compromising patient privacy as well as screening for diseases in low-data regime.

\section{Conclusion}
In this work, we share our practical experience in development to deployment of AI-assisted telehealth and screening workflows for vision- and hearing-threatening diseases in resource-constrained settings.
Our findings highlight real-world challenges that go beyond model accuracy: specifically, image acquisition constraints, time and usability bottlenecks, and early prototyping using public datasets and models for concrete interaction between AI and Clinical team. 

Our observations reinforce need for more systematic documentation of these end-to-end deployment pathways, especially in low-resource contexts with predominant paper-based high-volume workflows. 
By sharing these lessons learnt, we hope to inform and encourage future projects to bridge the gap between promising AI prototypes and robust screening solution to improve community health services.

\begin{credits}
\subsubsection{\ackname} This study is part of the Telehealth and AI-assisted screening Project funded
by DirectRelief provided to Hospital for Children, Eye, ENT and Rehabilitation Services (CHEERS), BP Eye Foundation. 
\subsubsection{\discintname}
The authors have no competing interests to declare that are
relevant to the content of this article.
\end{credits}
%
%
%
\bibliographystyle{splncs04}
\bibliography{sample}
\end{document}